\documentclass[letterpaper, 10 pt, conference]{ieeeconf/ieeeconf}
\IEEEoverridecommandlockouts
\usepackage{graphicx}
\usepackage{caption}
\usepackage{subcaption}
\usepackage{booktabs}
\usepackage{array}
\usepackage{multirow}
\usepackage{siunitx}
\usepackage{amsmath}
\usepackage{amssymb}
\usepackage[printonlyused]{acronym}
\usepackage[noEnd=true]{algpseudocodex}
\usepackage{algorithm}
\usepackage[hidelinks]{hyperref}
\acrodef{AD}{Automatic Differentiation}
\acrodef{BPTT}{Backpropagation Through Time}
\acrodef{CRBA}{Composite-Rigid-Body Algorithm}
\acrodef{DoF}{Degree of Freedom}
\acrodef{EoM}{Equation of Motion}
\acrodef{FoG}{First-order Gradient}
\acrodef{GD}{Gradient Descent}
\acrodef{LCP}{Linear Complementarity Problem}
\acrodef{LM}{Levenberg-Marquardt}
\acrodef{LSTM}{Long Short-Term Memory}
\acrodef{NCP}{Nonlinear Complementarity Problem}
\acrodef{PPO}{Proximal Policy Optimization}
\acrodef{RNEA}{Recursive Newton-Euler Algorithm}
\acrodef{RL}{Reinforcement Learning}
\acrodef{RSL}{Robotic Systems Lab}
\acrodef{SHAC}{Short-Horizon Actor-Critic}
\acrodef{ToI}{Time of Impact}
\acrodef{ZoG}{Zeroth-order Gradient}

\usepackage{balance}

\newcommand\Tstrut{\rule{0pt}{2.6ex}}         
\newcommand\Bstrut{\rule[-0.9ex]{0pt}{0pt}}   

\title{\LARGE \bf
DiffSim2Real: Deploying Quadrupedal Locomotion \\ Policies Purely Trained in Differentiable Simulation
}
\author{Joshua Bagajo$^{*,1}$, Clemens Schwarke$^{*,1}$, Victor Klemm$^{1}$, Ignat Georgiev$^{2,3}$,\\ Jean-Pierre Sleiman$^{1,3}$, Jesus Tordesillas$^{1,4}$, Animesh Garg$^{2}$, and Marco Hutter$^{1}$
\thanks{
${}^{*}$Shared 1st authorship.
${}^{1}$Robotic Systems Lab, ETH Zürich, Switzerland.
${}^{2}$Georgia Institute of Technology, United States.
${}^{3}$The AI Institute, United States.
${}^{4}$Institute for Research in Technology, ICAI School of Engineering, Comillas Pontifical University, Spain. \newline
CoRL 2024 Workshop 'Differentiable Optimization Everywhere'
}
}

\begin{document}
\maketitle
\thispagestyle{empty}
\pagestyle{empty}


\begin{abstract}

Differentiable simulators provide analytic gradients, enabling more sample-efficient learning algorithms and paving the way for data intensive learning tasks such as learning from images. In this work, we demonstrate that locomotion policies trained with analytic gradients from a differentiable simulator can be successfully transferred to the real world. Typically, simulators that offer informative gradients lack the physical accuracy needed for sim-to-real transfer, and vice-versa. A key factor in our success is a smooth contact model that combines informative gradients with physical accuracy, ensuring effective transfer of learned behaviors. To the best of our knowledge, this is the first time a real quadrupedal robot is able to locomote after training exclusively in a differentiable simulation.

\end{abstract}

\section{Introduction and Approach}

The majority of \ac{RL} algorithms rely on \ac{ZoG} estimates during optimization, allowing the use of conventional physics simulators that are typically non-differentiable. However, differentiable simulators offer analytically computed \acp{FoG}, with lower variance~\cite{mohamed2020monte,wiedemann2022,zhang2024back} and therefore improved sample efficiency and asymptotic policy performance~\cite{Suh2022,georgiev2024adaptive}. Thus, leveraging \acp{FoG} offers the potential to learn from pixels~\cite{luo2024residual} or to learn policies for systems with many degrees of freedom~\cite{xu2022}. Unfortunately, contact interactions are often simulated in a discontinuous manner, making \ac{FoG}-based optimization challenging. Some simulators address this by using soft contact models, which are continuous and smooth but less physically accurate for typical locomotion problems compared to discontinuous hard contact models~\cite{gehring2014evaluation}. Additionally, penalty-based soft contact models often require smaller time steps and thus increase computational demand and lengthen gradient chains. Consequently, learning the contact-rich task of quadrupedal locomotion and transferring the learned behavior to the real world with either hard or penalty-based contact has not yet succeeded~\cite{schwarke2024learning, freeman2021, Degrave2019}. Instead, we adopt an analytically smooth contact model, introduced in our previous work~\cite{schwarke2024learning}, that provides a smoothed optimization surface while maintaining physical accuracy, combining the advantages of hard and soft contact. The contact model draws inspiration from the role of stochasticity in current learning frameworks, a key factor in the success of \ac{RL}~\cite{Suh2022a,Pang2023}. We then employ the \ac{SHAC} algorithm \cite{xu2022} that leverages \acp{FoG} to enhance learning efficiency over purely \ac{ZoG}-based algorithms such as~\ac{PPO} \cite{schulman2017proximal}. Finally, we demonstrate that locomotion policies learned with this approach successfully transfer to the real world.

\begin{figure}[t]
   \centering
   \includegraphics[width=\columnwidth]{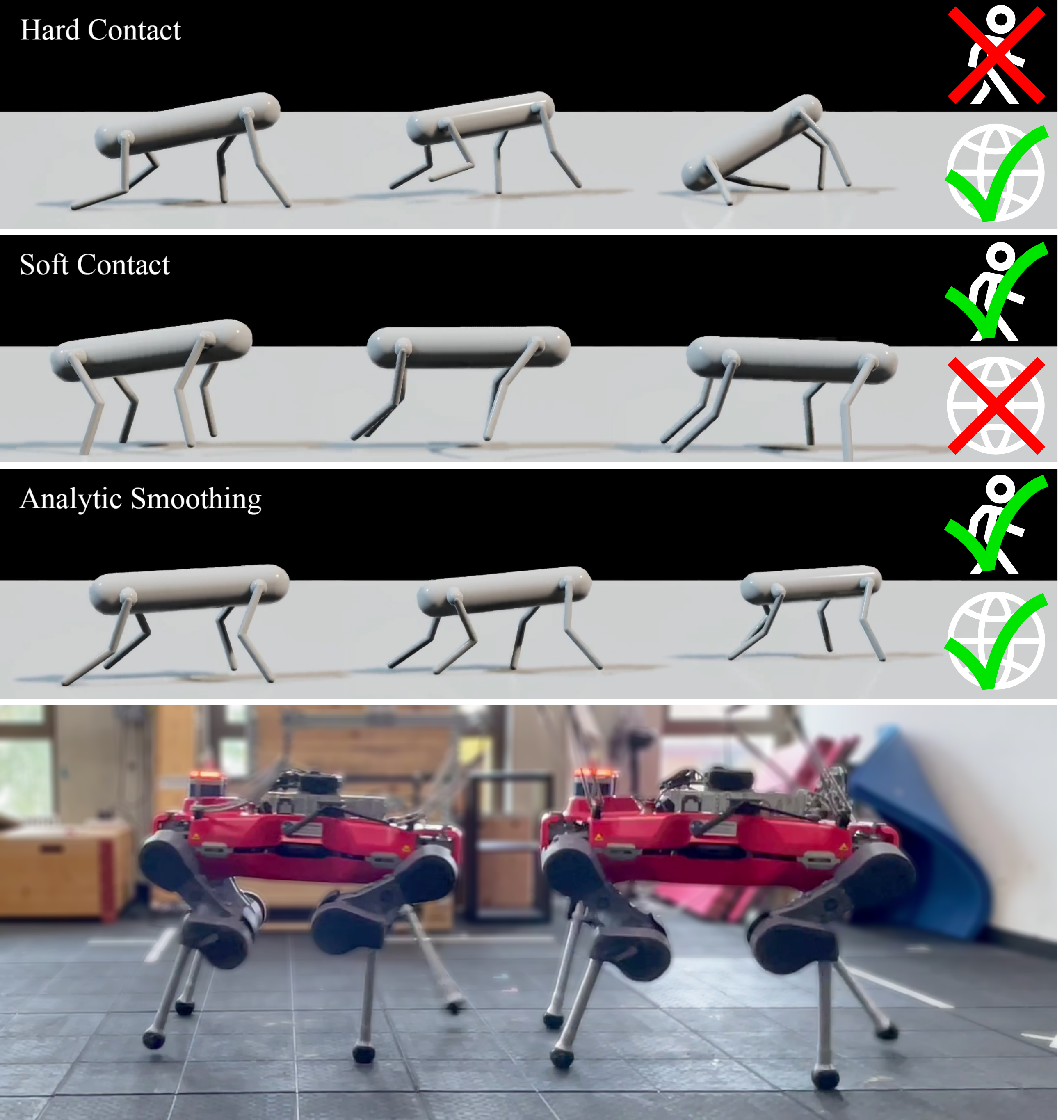}
   \caption{A quadrupedal robot learning to walk on flat terrain in a differentiable simulation. Policies trained with a \textit{hard contact model} follow unreasonable foothold patterns and do not learn to locomote robustly. Training with a \textit{soft contact model} results in stable locomotive gaits but the learned behaviors do not transfer to real hardware. Policies trained with an \textit{analytically smooth contact model} exhibit effective and stable locomotive gaits and transfer to the real world. Video: {\href{https://youtu.be/2wZmmUyqUQM}{https://youtu.be/2wZmmUyqUQM}}.}
   \label{fig:title}
   \vspace{-0.4cm}
\end{figure}

Previous attempts were confined to simulation. The first locomotion policy purely trained in a differentiable simulator was presented in~\cite{Degrave2019}, but exhibited undesirable behaviors like front flips. Table~\ref{tab:diffsims} summarizes relevant simulators and their approaches to differentiation and contact modeling. Nimble~\cite{Werling2021} implements symbolic differentiation and solves a sparse \ac{LCP} to resolve contact, while DiffTaichi~\cite{Hu2019} uses impulse-based methods to avoid differentiating the \ac{LCP} of contact. Warp~\cite{warp2022} and Brax~\cite{freeman2021} leverage GPU acceleration for fast rigid-body simulations and support multiple contact models. Dojo~\cite{Howell2022} emphasizes physical accuracy but is limited by slower execution and lacks parallelism. At the time of writing, none of the current simulators offer parallelization combined with accurate dynamics and informative gradients to learn transferable locomotion behaviors. A sim-to-real transfer for quadrupedal locomotion policies learned using \acp{FoG} was only achieved with a second non-differentiable simulator to ensure accurate physics~\cite{song2024learning}. In this work, we extended Warp with custom physics to benefit from GPU parallelization.

\begin{table}[htbp]
\caption{Differentiable Rigid-Body Simulators}
\label{tab:diffsims}
\begin{center}
\begin{tabular}{l c c c}
\toprule
Name & Differentiation & Contact Modeling & Device\\
\hline
Nimble~\cite{Werling2021} & Symbolic & \acs{LCP} & CPU\Tstrut\\
DiffTaichi~\cite{Hu2019} & Automatic & Impulse-based & GPU\\
Warp~\cite{warp2022} & Automatic & XPBD~\cite{Macklin2016}, Soft & GPU\\
Brax~\cite{freeman2021} & Automatic & MuJoCo~\cite{Todorov2012MuJoCoAP}, PBD~\cite{muller2007position} & GPU\\
Dojo~\cite{Howell2022} & Symbolic & \acs{NCP} & CPU\Bstrut\\
\bottomrule
\end{tabular}
\end{center}
\vspace{-0.4cm}
\end{table}

\section{Contact Simulation}
\label{sec:contact}

\begin{figure}[htbp]
   \centering
   \includegraphics[width=\columnwidth]{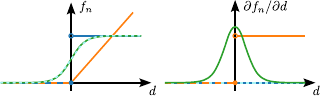}
   \caption{The normal contact force (left) and its gradient (right) with respect to the penetration depth between two contacting bodies. \textit{Hard contact (blue)} exhibits a discontinuity at $d=0$. Its analytical gradient is zero almost everywhere. \textit{Soft contact (orange)} is continuous but does not accurately model stiff contact without becoming unstable because the normal force is unbounded. \textit{Stochastically smoothing hard contact (cyan)} removes the discontinuity, but the \ac{FoG} gradient remains zero and thus uninformative. \textit{Analytically smoothing hard contact (green)} induces similar effects on the dynamics as stochastic smoothing, with the advantage of an informative \ac{FoG}.}
   \label{fig:analytic_smoothing}
   \vspace{-0.4cm}
\end{figure}

Our simulation is based on Moreau's time stepping scheme~\cite{moreau1988unilateral}. However, the Gauss-Seidel algorithm used to compute contact forces is slightly modified from implementations such as~\cite{Carius2018} to smooth the originally hard contact model. Contact forces are scaled by a sigmoid function that depends on the distance between potentially contacting bodies. For more details on the simulation, we refer to~\cite{schwarke2024learning}. The analytically smooth contact model has several advantages over hard and soft contact models. First, it smooths the discontinuities of hard contact. While stochastic smoothing would have similar effects on the dynamics, gradients would still remain uninformative \acp{FoG} as explained in Fig.~\ref{fig:analytic_smoothing}. Second, the contact model remains stable for larger simulation time steps compared to traditional soft contact models. Lastly, the similarity to stochastically smoothed dynamics suggests that the hard contact case is implicitly within the domain of the analytically smooth contact model, promising successful transfer to hard contact or the real world.

\section{Sim-to-Real Transfer}
\label{sec:s2r}

To transfer locomotion policies learned in our differentiable simulation to ANYbotics' ANYmal D robot, we first align our learning setup with~\cite{orbit2023, rudin2022learning}, which have demonstrated successful sim-to-real transfer. To test the validity of the dynamics of our simulation, we train policies with \ac{PPO} in our simulator and transfer them to IsaacSim, a hard contact simulation used in~\cite{orbit2023, rudin2022learning}. After ensuring that the learned behaviors successfully transfer, we progress to training policies with \ac{SHAC}, making use of the \acp{FoG} computed by the simulator. However, the learning setup designed for \ac{PPO} does not immediately lead to the desired behaviors with our method. Instead, our method requires a simplified inertia model (only diagonal inertia components with lower magnitudes) to find a reasonable locomotion policy. The reward formulation needs adaptation as well. We find that combining rewards from~\cite{schwarke2024learning} with rewards from~\cite{rudin2022learning} that allow for differentiation results in successful learning.

Key elements for sim-to-real transfer, according to~\cite{rudin2022learning}, are domain randomization and the integration of a learned actuator model. Initially, we adopt the domain randomization method from~\cite{rudin2022learning}, though the required extent of randomization for our approach remains to be determined. While domain randomization helps to close the sim-to-real gap and smooths out local minima in the learning objective, we observe that higher levels of randomization lead to slower convergence during training. Incorporating an actuator model, which typically contains a history or memory architecture, would significantly increase the complexity of the computational graph. Instead, we implement a PD-controller with velocity-based torque saturation, using system identification-derived parameters~\cite{bjelonic2024sim2real}. This approach provides efficient gradient propagation and achieves performance comparable to the learned actuator model in the relevant actuation domain, without adding unnecessary complexity to the computational graph.

\section{Results and Limitations}
\label{sec:results}

In previous work~\cite{schwarke2024learning}, we found that common soft and hard contact models do not lead to transferable locomotion policies, as shown in Fig.~\ref{fig:title}. Our introduction of analytic smoothing enabled smooth gaits that successfully transferred to hard contact simulation. In this work, we further show that policies learned in a differentiable simulator also transfer effectively to real-world environments. However, learning locomotion with \acp{FoG} is sensitive to physical parameters and the reward function, and our approach has not yet surpassed state-of-the-art \ac{RL} policies in terms of locomotion behavior. Nevertheless, learning with \ac{SHAC} requires significantly fewer samples---over an order of magnitude less---compared to \ac{PPO}. Although we presented a proof of concept in this preliminary work, an in-depth analysis will be necessary in future research. Furthermore, we plan to introduce roughness to the terrain to enhance the locomotion behavior, as the current flat terrain limits stepping height and robustness.

\balance
\bibliographystyle{IEEEtran}
\bibliography{bibliography/references}

\end{document}